# A Co-Training Semi-Supervised Framework Using Faster R-CNN and YOLO Networks for Object Detection in Densely Packed Retail Images

Hossein Yazdanjouei [1], Arash Mansouri [2] and Mohammad Shokouhifar [3,*]

[1] *Department of Computer Science, Khazar University, Baku AZ1096, Azerbaijan*
[2] *Department of Computer Engineering, Politecnico di Milano, Milano 20133, Italy*
[3] *Institute of Research and Development, Duy Tan University, Da Nang 550000, Vietnam*

***Abstract***: This study proposes a semi-supervised co-training framework for object detection in densely packed retail environments, where limited labeled data and complex conditions pose major challenges. The framework combines Faster R-CNN (utilizing a ResNet backbone) for precise localization with YOLO (employing a Darknet backbone) for global context, enabling mutual pseudo-label exchange that improves accuracy in scenes with occlusion and overlapping objects. To strengthen classification, it employs an ensemble of XGBoost, Random Forest, and SVM, utilizing diverse feature representations for higher robustness. Hyperparameters are optimized using a metaheuristic-driven algorithm, enhancing precision and efficiency across models. By minimizing reliance on manual labeling, the approach reduces annotation costs and adapts effectively to frequent product and layout changes common in retail. Experiments on the SKU-110k dataset demonstrate strong performance, highlighting the scalability and practicality of the proposed framework for real-world retail applications such as automated inventory tracking, product monitoring, and checkout systems.

***Keywords:*** *Retail object detection; Densely packed scenes; Semi-supervised learning; Co-training method; Faster R-CNN; Metaheuristic optimization; YOLO integration.*

**\* Correspondence:** Mohammad Shokouhifar
Email: mohammadshokouhifar@duytan.edu.vn, ORCID: 0000-0001-7370-4760.

**Email addresses:**

Hossein Yazdanjouei: h.yazdanjouei@gmail.com  (ORCID: 0000-0002-3513-3781)
Arash Mansouri: arash@pxp.ai  (ORCID: 0000-0003-3431-2195)
Mohammad Shokouhifar: mohammadshokouhifar@duytan.edu.vn (ORCID: 0000-0001-7370-4760)

# 1. INTRODUCTION

Detecting objects in densely packed retail environments has become essential due to the increasing demand for automation in inventory management, product recognition, and efficient checkout processes in modern retail. Retail shelves, often stacked with numerous small, visually similar items, create a uniquely challenging environment for object detection algorithms. Efficient and precise recognition of these items helps optimize retail operations, such as stock replenishment, and improves customer experience by reducing wait times at checkout. In retail settings, where products frequently vary in packaging and appearance, accurate object detection is critical for ensuring consistency across tasks, from pricing to inventory updates. Consequently, designing robust object detection solutions that can handle this complexity has garnered significant interest in both research and industry [1]. While barcodes and RFID systems have been traditionally employed for object detection in retail, they face substantial limitations in handling densely packed product settings. Barcode scanning typically requires manual operation, making it unsuitable for continuous monitoring, while RFID requires individual tags for each item, which can significantly increase costs and operational complexity. Also, these methods lack the flexibility to adapt seamlessly to dynamic retail environments, where products and layouts frequently change [2].

In contrast, image processing approaches offer a scalable and non-intrusive alternative by recognizing items based solely on their visual attributes. However, detecting objects in densely packed retail settings introduces specific challenges, primarily due to significant occlusions and overlaps among items that complicate both accurate identification and localization. Traditional object detection models often struggle in these conditions, frequently yielding overlapping bounding boxes, misclassifications, and increased false positives. Deep learning models, such as Faster R-CNN [3] and YOLO [4], have been proposed in the literature to tackle these challenges; however, they require vast amounts of labeled data to achieve high accuracy in such complex scenarios. Creating such extensive labeled datasets is time-consuming, labor-intensive, and costly, especially for environments with high object density, like retail shelves [5]. Additionally, in retail, product labels and designs are frequently updated, as new products are introduced regularly. Continuous manual updates to labels are impractical, and models trained on static, labeled datasets risk losing relevance over time, as they fail to account for these evolving product displays.

Semi-supervised learning (SSL) has emerged as a promising approach to address these limitations by using both labeled and unlabeled data to enhance model performance. In the context of densely packed scenes, SSL allows models to draw on the extensive amounts of unlabeled retail images, gradually refining their detection capabilities and gaining robustness to occlusions and object overlaps [6]. SSL can enable object detection models to generalize better in retail settings, where acquiring exhaustive labels is resource-intensive. By learning from a smaller set of labeled examples alongside a large pool of unlabeled data, SSL reduces reliance on manual annotation while improving model accuracy and robustness. This is particularly advantageous in retail, where SSL facilitates the adaptation of detection models to new product designs without requiring constant manual labeling, ensuring that detection remains accurate and consistent in evolving store layouts.

Within the broader semi-supervised learning field, several methodologies exist, including self-training, teacher-student, and co-training, each offering unique benefits [7]. In self-training, a single model iteratively labels its own predictions as pseudo-labels, but this approach can lead to accumulated errors if initial predictions are inaccurate. The teacher-student framework addresses this limitation by introducing a more knowledgeable "teacher" model to guide a "student" model, thereby generating more reliable pseudo-labels. However, the effectiveness of this framework still depends significantly on the initial performance of the teacher model. Co-training, in contrast, employs two independent models, each providing pseudo-labels to the other on unlabeled data, allowing for cross-reinforcement and correction. By employing the diversity between two different models, co-training reduces biases specific to each model and increases detection accuracy, making it particularly suitable for detecting densely packed retail objects. The collaborative learning mechanism of co-training has proven to yield more robust predictions, as each model benefits from the unique perspectives and detection capabilities of its counterpart, ultimately achieving superior results in complex, densely packed scenes.

This study presents a co-training semi-supervised approach for object detection in densely packed retail scenes, demonstrated using the SKU-110k dataset, which underscores the limitations of relying solely on labeled data in complex retail settings. The framework integrates two models, Faster R-CNN with a ResNet backbone and YOLO with a Darknet backbone, working collaboratively to enhance detection accuracy through mutual pseudo-label generation, allowing each model to utilize the other's strengths. Faster R-CNN excels in precise localization and boundary definition, while YOLO's capacity for global context understanding enables effective detection in cluttered scenes with overlapping items. After feature extraction, classification is achieved through an ensemble of XGBoost (XGB), Random Forest (RF), and Support Vector Machine (SVM) classifiers, each adding unique capabilities for robust classification within high-density images. Hyperparameter tuning is performed using a metaheuristic-driven algorithm, optimizing parameters across the ensemble and detection models to enhance performance, accuracy, and adaptability to varied retail conditions. This co-training approach thus combines precise localization, global context capture, and robust classification to effectively address the challenges of detecting items in densely packed retail environments, ultimately achieving enhanced detection performance on the SKU-110k dataset [8] and presenting a practical solution for complex retail environments.

The remaining sections of the paper are organized as follows: Section 2 provides a comprehensive review of recent influential studies utilizing the SKU-110k dataset for object detection in dense retail environments. Section 3 outlines the proposed methodology, including a detailed explanation of the co-training approach and the integration of metaheuristic-driven optimization for hyperparameter tuning. Section 4 presents and discusses the experimental results achieved using the proposed approach, with a focus on assessing model performance and effectiveness in densely packed retail scenarios. Finally, Section 5 concludes the paper by summarizing key findings and offering recommendations for future research directions in this field.

## 2. LITERATURE REVIEW

As mentioned above, the SKU-110K dataset has emerged as a critical resource for advancing object detection techniques, particularly in densely packed retail environments where traditional models struggle with precision and efficiency. This section provides a review of key studies that have employed the SKU-110K dataset to enhance detection precision, robustness, and scalability across complex scenes and crowded spaces.

Goldman et al. [8] introduced the SKU-110K dataset, a large-scale dataset specifically designed for densely packed object detection in retail environments. They proposed a novel detection framework that integrated a Soft-IoU layer to refine region proposals, addressing the difficulty of detecting overlapping objects. Additionally, their EM-Merger unit helped resolve detection overlaps by leveraging the Intersection-over-Union (IoU) scores, resulting in more accurate object localization. Evaluated across SKU-110K, CARPK, and PUCPR+ datasets, their model significantly reduced false positives, outperforming existing models, particularly in environments where high object density makes detection challenging. Their method set a new benchmark in crowded retail settings, with substantial improvements in both detection precision and recall.

Building on the challenges posed by oriented and densely packed objects, Pan et al. [9] developed the Dynamic Refinement Network (DRN), which introduced two key innovations: the Feature Selection Module (FSM), which dynamically adjusts neuron receptive fields based on object orientation, and the Dynamic Refinement Head (DRH), which refines detection predictions by considering object-aware factors. The model was rigorously evaluated on several challenging datasets, including SKU-110K, SKU-110K-R, DOTA, and HRSC2016, showing significant improvements over baseline models such as Faster R-CNN and YOLOv3. By improving localization precision, DRN excelled in detecting oriented objects in both retail environments and complex aerial imagery, where object orientation and dense arrangements frequently hinder traditional detection models.

Ye et al. [6] approached the problem of limited labeled data in densely packed object detection by proposing a semi-supervised learning framework. Their framework employed a teacher-student model and introduced two novel loss functions (IoU-aware consistency loss and proposal consistency loss) to improve the consistency between the teacher's predictions and the student's proposals. This method allowed for the effective use of unlabeled data, significantly enhancing detection performance in environments with small and overlapping objects. Tested on the RebarDSC and SKU-110K datasets, the framework demonstrated substantial improvements over existing semi-supervised methods. This work is particularly impactful in domains where labeled data is scarce but critical for accurate detection, such as densely packed retail scenes.

In their work on SKU-level product recognition, Pietrini et al. [10] proposed a deep learning-based system that employs a two-network pipeline designed to solve the challenge of identifying products in dynamic and high-density retail environments. The first network detects products using the SKU-110K dataset, which serves as the foundation for detecting products on shelves in crowded conditions. The second network generates embedding vectors that represent critical product features, such as shape, color, and size. These vectors are then compared using cosine similarity against a custom EAN-labeled dataset. This two-stage pipeline successfully facilitated precise product identification, showcasing the system's scalability and robustness in real-world retail environments where rapid and accurate SKU recognition is essential for applications like automated checkout and inventory tracking.

Similarly, Gothai et al. [11] designed a two-stage pipeline for product recognition in grocery retail environments using the YOLOv5 detection model combined with Naive Bayes Similarity Search and Fisher Kernels for product recognition. Their approach focused on enhancing feature extraction by using color, shape, and size features to distinguish visually similar products. Tested on the SKU-110K dataset, the pipeline exhibited substantial improvements in accuracy, precision, and recall, especially when recognizing products with similar packaging designs. By incorporating Fisher Kernels into the feature extraction process, the model improved high-dimensional feature representation, making it highly effective for real-world applications such as automated checkout systems and inventory management.

To address the challenge of detecting densely packed objects, Xu et al. [12] proposed a method combining Cascade R-CNN with Feature Pyramid Networks (FPN). Their model incorporated Complete Intersection over Union (CIOU) loss and balanced L1 loss to improve the accuracy of region proposals and handle overlapping objects more effectively. Tested on the SKU-110K dataset, their method achieved an mAP of 0.413, which marked a significant improvement over previous models in terms of detection accuracy in dense retail environments. The use of CIOU loss allowed for more refined bounding box predictions, making the model particularly adept at handling the complexity of densely packed product arrangements in retail settings.

Cho et al. [13] developed a detection method focused on improving the accuracy of object detection in densely packed scenes by combining Advanced Weighted Hausdorff Distance (AWHD) with a Hard Negative-Aware Anchor Attention (HNAA) mechanism. The AWHD helped improve object center estimation by mitigating misalignments between ground truth and predicted boxes, while the HNAA penalized hard-to-detect negative anchors that are often mistaken for true positives in densely packed environments. Tested on SKU-110K, WebMarket, Holoselecta, and CAPG-GP datasets, their model demonstrated superior performance, particularly in identifying small, overlapping objects in crowded retail settings.

An alternative approach was taken by Hong et al. [14], who developed an IoU-aware Feature Fusion R-CNN model to handle multi-scale objects in densely packed scenes. By integrating the IoU metric directly into the feature extraction process, their model enhanced localization precision, especially for small and densely packed objects. Evaluated on the SKU-110K dataset, the model outperformed baseline approaches like Faster R-CNN with FPN, achieving better precision and recall while maintaining a simplified model architecture that is more computationally efficient for real-time detection scenarios.

In the context of transformer-based architectures, Dai and Liu [15] presented DeCo-DETR, a novel framework specifically designed for commodity detection in retail environments with densely packed products. Their model introduced an adaptive positional prior generator and a density-map-guided ranker assignment strategy to prevent incorrect associations between predicted and ground truth boxes. Evaluated on the SKU-110K dataset, DeCo-DETR achieved state-of-the-art performance, excelling in terms of precision and recall, and proving particularly effective in handling small and overlapping objects in complex retail scenarios.

Enhancing object detection in densely packed scenes, Zhong et al. [16] integrated a Transformer-based head with YOLOv5. This approach introduced a self-attention mechanism within the transformer head, enabling the model to effectively capture long-term dependencies between objects—a critical feature for differentiating closely spaced items. The model also incorporated an EM-Merger unit designed to eliminate redundant detections, which led to more precise predictions for overlapping objects. Tested on the SKU-110K and RebarDSC datasets, this method demonstrated notable improvements in precision and recall, showcasing the strength of combining transformer architectures with established detection methods in retail and industrial environments.

Vasanthi and Mohan [17] developed a transformer-based model focused on detecting x-small and densely packed objects. Their approach addressed the issue of feature misalignment between anchor boxes and convolutional layers, using an auto-anchor module and Multi-Head Self-Attention (MHSA) to enhance detection accuracy. By dynamically adjusting anchor scales and extracting more detailed feature maps, the model showed superior performance in terms of precision, recall, and mAP when evaluated on VOC, VisDrone, and SKU-110K datasets. This model proves to be highly valuable for dense detection tasks, particularly in challenging environments like retail and aerial imagery.

To address the complexity of handling large supermarket datasets, Strohmayer and Kampel [18] introduced a real-time product recognition pipeline optimized for mobile devices. The pipeline employs the Global Trade Item Number (GTIN) system, enabling it to efficiently handle dynamic supermarket inventories. Tested on the SKU110K and R6k datasets, it successfully recognized almost 6,000 products with an impressive inference time of 121ms per image on a Google Pixel 6, achieving approximately 8.3 frames per second (fps). This system demonstrates its real-world applicability for rapid product recognition in retail.

Improving SKU recognition in unmanned retail stores was the focus of Wang et al. [19], who developed a robust residual network architecture. The framework includes a boundary regression neural network for detecting bounding boxes and incorporates weighted non-maximum suppression (WNMS) to further enhance detection precision. Their approach, evaluated on the SKU110K and RPC datasets, yielded significant gains in both accuracy and computational efficiency, making it highly suitable for real-time SKU checkout systems in modern retail.

A hybrid method combining traditional computer vision techniques and deep learning was proposed by Melek et al. [20] to improve grocery product recognition. The method integrates Aggregate Channel Features (ACF) and Single-Shot Detector (SSD) for product detection, along with SURF, BRISK, and ORB for feature extraction, to enhance recognition accuracy. This approach was tested on the SKU110K and Grocery Products datasets, achieving notable improvements in accuracy and scalability, making it applicable for automated checkout and inventory management in retail environments.

Wang et al. [21] introduced a novel approach to densely packed object detection by developing DeIoU, a loss function aimed at reducing overlap between predicted boxes. Combined with a one-to-many (O2M) label matching strategy, this method significantly enhanced detection precision and object shape prediction. Evaluations on datasets such as SKU-110K, CrowdHuman, MS COCO 2017, and DIOR showed an average improvement of 1.3 AP and 1.8 MR-2, demonstrating its effectiveness in handling densely packed object detection tasks with high precision.

Finally, the paper by Zhao et al. [5] introduced FSSLOD (Federated Semi-Supervised Learning framework for Object Detection), targeting object detection in densely packed industrial settings. Addressing data privacy concerns and the high costs of manual labeling, the study utilized a federated learning approach combined with a semi-supervised learning structure. This framework employed a teacher-student model on the client side, which included a consistency loss mechanism to ensure alignment between teacher and student network outputs, and an elastic update mechanism to handle data distribution disparities across multiple institutions. Evaluations conducted on real-world object detection datasets demonstrated that FSSLOD significantly improved detection accuracy and model robustness while ensuring data privacy, presenting a valuable solution for densely packed object detection in industrial applications.

## 2.1. Our contributions against previous approaches

Previous studies have advanced retail object detection in dense shelf settings but remain limited by their heavy reliance on costly labeled datasets and single-model semi-supervised methods that struggle with overlapping and occluded products. To overcome these challenges, this study proposes a co-training semi-supervised framework that combines Faster R-CNN's precise localization with YOLO's global context, enabling mutual pseudo-labeling to improve accuracy in complex scenes. By reducing dependence on manual labeling, the approach adapts more easily to frequent product and layout changes while its dual-model design mitigates model-specific biases. Metaheuristic-driven hyperparameter tuning further boosts accuracy and robustness, making the framework a scalable and practical solution for real-world retail applications such as inventory tracking and automated checkout.

## 3. PROPOSED METHODOLOGY

This section provides a detailed overview of the proposed approach developed to tackle the unique challenges of object detection in densely packed retail environments, exemplified by the SKU-110k dataset. As real-world retail environments are dynamic, with frequent updates to product displays and layouts, our approach aims to reduce dependency on extensive manual labeling while maintaining high detection accuracy for object detection in densely packed retail scenes. Figure 1 demonstrates the overall architecture of the proposed approach.

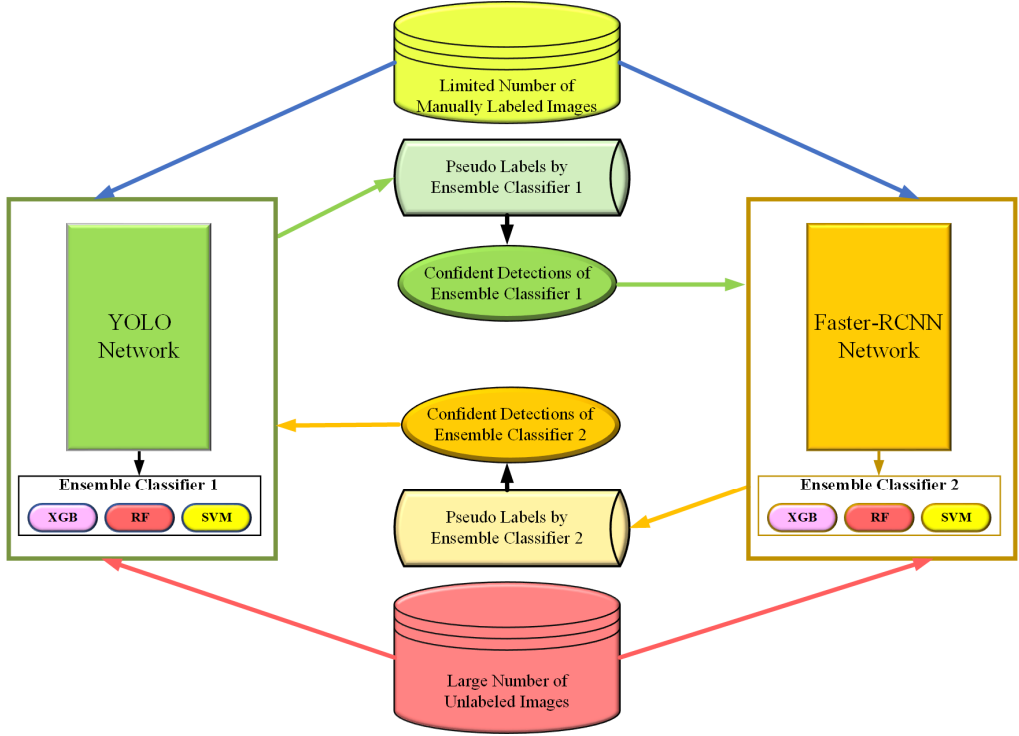

Fig. 1. The overall flowchart of the proposed methodology

## 3.1. Input Dataset and Data Preparation

Before The SKU-110k dataset, introduced by Goldman et al. [8], is designed to address the challenge of object detection in densely packed retail environments, such as grocery shelves where products are placed closely together, often with minimal spacing and numerous overlapping items. It consists of over 11,000 images across a variety of retail products, capturing both high object density and complex visual patterns. These characteristics make it particularly valuable for training and evaluating object detection models that can navigate occlusions and overlapping objects effectively.

For this study, 2,000 labeled images and 8,000 unlabeled images were selected from the SKU-110k dataset to support a semi-supervised approach. The presence of unlabeled data allows for extensive pseudo-label generation, enhancing the models' ability to learn without increasing annotation costs. The labeled data was divided into three subsets to optimize the training and evaluation process: 70% for training, 10% for validation, and 20% for testing. This 70-10-20 split allows the training set to support the initial supervised learning phase, the validation set to guide hyperparameter tuning, and the test set to provide a final performance evaluation of the model's generalization ability.

## 3.2. Model Architecture and Feature Extraction

In this framework, two distinct object detection models are employed to utilize their complementary strengths in feature extraction and detection accuracy.

- Faster R-CNN with ResNet Backbone
- YOLO with Darknet Backbone.

Both models process the same input data from the SKU-110k dataset, yet each emphasizes distinct aspects of feature extraction: Faster R-CNN's architecture is optimized for high localization accuracy, which is crucial for defining tight and precise object boundaries, while YOLO's design is centered around capturing spatial context and managing occlusions through its global view. This integration of detailed and contextual feature extraction achieves a well-rounded approach to detection, where Faster R-CNN ensures precision in dense scenes, and YOLO enables a broader context-aware perspective. This complementary pairing results in a balanced architecture that enhances detection accuracy, robustness, and adaptability in densely packed retail environments, ultimately contributing to more effective retail object recognition.

### 3.3. Ensemble Classifiers for Feature Classification

Following feature extraction, each model (Faster R-CNN and YOLO) passes its output to a series of ensemble classifiers that include XGB, RF, and SVM. Ensemble learning is a technique that combines multiple models to make predictions, offering higher accuracy, improved robustness, and better generalization than relying on a single model alone [22, 23]. Each of the utilized classifiers within the ensemble model (i.e., XGB, RF, and SVM) contributes unique strengths to the classification of object features. Each deep model (Faster R-CNN and YOLO) generates feature vectors that are classified by the ensemble, producing initial predictions. These initial predictions are used to assign pseudo-labels to unlabeled data within the training set, a core component of the semi-supervised co-training approach.

### 3.4. Pseudo-Label Generation and Exchange

In the co-training framework, pseudo-labels generated by the ensemble classifiers from one model (e.g., Faster R-CNN) are shared with the other model (e.g., YOLO) as additional training data. This mutual pseudo-label exchange creates a feedback loop where each model can reinforce the strengths of the other. The process is iterative, allowing both models to refine their predictions based on the pseudo-labels provided by their counterpart. This iterative exchange of pseudo-labels helps both models improve their detection performance by focusing on both precise localization (Faster R-CNN) and global context (YOLO). For example, in the first iteration, Faster R-CNN may provide pseudo-labels with high localization accuracy. YOLO can then use these pseudo-labels to improve its detection of occluded or overlapping objects. In the subsequent iteration, YOLO provides pseudo-labels that incorporate its understanding of global context, which Faster R-CNN can use to refine its predictions, especially for objects that are partially occluded. By iteratively exchanging pseudo-labels, the co-training approach achieves more accurate and reliable predictions across the entire dataset.

### 3.5. Metaheuristic-driven Hyperparameter Tuning

To optimize detection performance, hyperparameter tuning is conducted using a metaheuristic-driven optimization algorithm. Generally, metaheuristics are general-purpose high-level random-based optimization techniques which have shown promise performance across diverse engineering optimization problems, especially in the case of hyperparameter tuning domains [24-35]. This hyperparameter tuning process is applied in the initial phase of the co-training process, where the models are trained using only the labeled images, to optimize the parameters for the ensemble classifiers (XGB, RF, and SVM) as well as the Faster R-CNN and YOLO networks by maximizing mean Average Precision (mAP) on the validation set.

In object detection, mAP measures the model's precision across multiple Intersection over Union (IoU) thresholds, providing a comprehensive metric for evaluating both localization and classification accuracy. Key hyperparameters, such as learning rates, maximum depth, regularization coefficients, and tree count for XGB and RF, are fine-tuned to enhance model performance. This tuning process is critical for ensuring that both the detection models and ensemble classifiers operate at their full potential, effectively adapting to the dense and complex retail scenes represented in the SKU-110k dataset.

By optimizing based on mAP scores, the metaheuristic-driven optimization algorithm helps the models and classifiers generalize effectively across varied retail scenes, maintaining accuracy and robustness even in changing retail environments. Figure (2) shows a feasible representation for each solution to find the optimized hyperparameters by the metaheuristic algorithm. In this Figure, $LR_{XGB}$, $D_{XGB}$, $RC_{XGB}$, and $NT_{XGB}$ represent the learning rate, maximum depth, regularization coefficient, and number of trees for XGBoost; $D_{RF}$ and $NT_{RF}$ denote the depth and number of trees for Random Forest; $C_{SVM}$, $K_{SVM}$, and $G_{SVM}$ specify the regularization, kernel, and gamma for SVM; and $EP_{YOLO}$, $CT_{YOLO}$, $IOU_{YOLO}$, $BS_{YOLO}$, $LR_{YOLO}$, $EP_{RCNN}$, $CT_{RCNN}$, $IOU_{RCNN}$, $BS_{RCNN}$ and $AS_{RCNN}$ correspond to the number of epochs, confidence threshold, IoU threshold, batch size, learning rate and anchor scales for YOLO and Faster R-CNN, respectively.

| $LR_{XGB}$ | $D_{XGB}$ | $RC_{XGB}$ | $NT_{XGB}$ | $D_{RF}$ | $NT_{RF}$ | $C_{SVM}$ | $K_{SVM}$ | $G_{SVM}$ | $EP_{YOLO}$ |
|---|---|---|---|---|---|---|---|---|---|
| $CT_{YOLO}$ | $IOU_{YOLO}$ | $BS_{YOLO}$ | $LR_{YOLO}$ | $EP_{RCNN}$ | $CT_{RCNN}$ | $IOU_{RCNN}$ | $BS_{RCNN}$ | $LR_{RCNN}$ | $AS_{RCNN}$ |

Fig. 2. A feasible solution representation for hyperparameters tuning.

## 4. NUMERICAL EXPERIMENTS

This section offers a detailed evaluation of the proposed co-training semi-supervised model, benchmarked against several prominent object detection frameworks commonly applied to densely packed retail images. The assessment is conducted on the SKU-110k dataset, using consistent Python-based implementations for all methods. The experiments were run on a Windows system featuring an Intel Core i7-14700K CPU, an NVIDIA RTX 4080 GPU, and 32GB of RAM. Through this comparative analysis, we aim to provide clear insights into the effectiveness of our approach and its relative strengths compared to other models in densely packed retail detection tasks.

Figure 3 illustrates the detection results produced by our framework, displaying sample images from the SKU-110k dataset that highlight the method's effectiveness in identifying densely packed retail items. The detection outcomes suggest that our approach enables the model to accurately recognize and distinguish objects even within complex, cluttered scenes. These visualizations underscore the model's capacity to extract and learn refined features from unlabeled data, significantly improving detection accuracy. The results showcase the advantages of co-training semi-supervised learning in enhancing the model's feature extraction capabilities, making it particularly effective for dense object detection in challenging retail environments.

Evaluation on the SKU-110k dataset shows that our Co-training + Ensemble model achieves a mAP of 0.596, AP.75 of 0.663, and AR300 of 0.627, surpassing other rival models. These results validate our model's applicability to real-world retail scenarios, such as inventory tracking and automated checkout systems for retail stores, where precise and adaptable detection is essential.

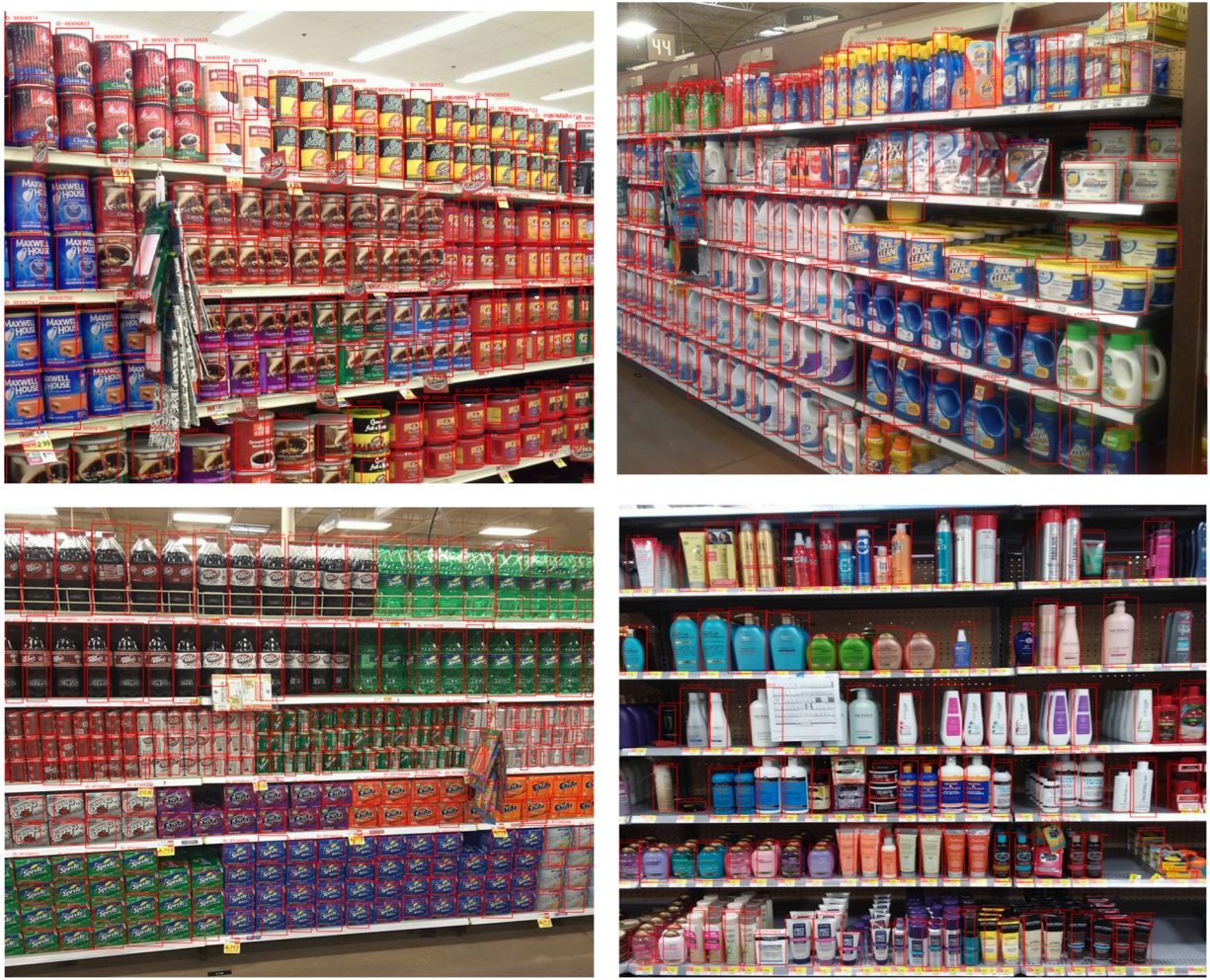

Fig. 3.  Sample images showing object detection results of the proposed algorithm applied to the SKU-110k dataset.

## 5. CONCLUSION

This study presents a novel framework (co-training + ensemble) tailored for object detection in densely packed retail settings. By combining Faster R-CNN for precise localization with YOLO for spatial context, our approach effectively handles the challenges posed by overlapping and occluded objects. The model employs semi-supervised learning, with each component generating pseudo-labels to support the other and significantly reducing the dependence on labeled data, which is a critical advantage given that manual labeling is both time-consuming and labor-intensive. Additionally, utilizing the ensemble classification fine-tuned by the metaheuristic-driven optimization, enhances detection accuracy and makes the model more resilient to changes in products' sizes and layouts by integrating the classification capabilities of XGB, RF, and SVM. Future work could explore integrating transformer-based architectures, which excel in capturing spatial relationships and fine-grained details, potentially enhancing detection accuracy in densely packed scenes. Another promising direction could involve online training mechanisms that allow the model to adapt to new products or layout changes without extensive retraining.

## REFERENCES

[1] Melek, C. G., Sönmez, E. B., & Varlı, S. (2024). Datasets and methods of product recognition on grocery shelf images using computer vision and machine learning approaches: An exhaustive literature review. *Engineering Applications of Artificial Intelligence*, *133*, 108452.

[2] Guimarães, V., Nascimento, J., Viana, P., & Carvalho, P. (2023). A review of recent advances and challenges in grocery label detection and recognition. *Applied Sciences*, *13*(5), 2871.

[3] Ren, S., He, K., Girshick, R., & Sun, J. (2016). Faster R-CNN: Towards real-time object detection with region proposal networks. *IEEE transactions on pattern analysis and machine intelligence*, *39*(6), 1137-1149.

[4] Redmon, J., Divvala, S., Girshick, R., & Farhadi, A. (2016). You only look once: Unified, real-time object detection. In *Proceedings of the IEEE conference on computer vision and pattern recognition* (pp. 779-788).

[5] Zhao, C., Gao, Z., Bao, S., & Xiao, K. (2024). Toward industrial densely packed object detection: a federated semi-supervised learning approach. *IEEE Internet of Things Journal*.

[6] Ye, C., Zhang, H., Xu, X., Cai, W., Qin, J., & Choi, K. S. (2021, August). Object Detection in Densely Packed Scenes via Semi-Supervised Learning with Dual Consistency. In *IJCAI* (pp. 1245-1251).

[7] Ning, X., Wang, X., Xu, S., Cai, W., Zhang, L., Yu, L., & Li, W. (2023). A review of research on co-training. *Concurrency and computation: practice and experience*, *35*(18), e6276.

[8] Goldman, E., Herzig, R., Eisenschtat, A., Goldberger, J., & Hassner, T. (2019). Precise detection in densely packed scenes. In *Proceedings of the IEEE/CVF conference on computer vision and pattern recognition* (pp. 5227-5236).

[9] Pan, X., Ren, Y., Sheng, K., Dong, W., Yuan, H., Guo, X., ... & Xu, C. (2020). Dynamic refinement network for oriented and densely packed object detection. In *Proceedings of the IEEE/CVF conference on computer vision and pattern recognition* (pp. 11207-11216).

[10] Pietrini, R., Rossi, L., Mancini, A., Zingaretti, P., Frontoni, E., & Paolanti, M. (2022, May). A deep learning-based system for product recognition in intelligent retail environment. In *International Conference on Image Analysis and Processing* (pp. 371-382). Cham: Springer International Publishing.

[11] Gothai, E., Bhatia, S., Alabdali, A., Sharma, D., Kondamudi, B. R., & Dadheech, P. (2022). Design features of grocery product recognition using deep learning. *Intelligent Automation and Soft Computing*, *34*(2), 1231-1246.

[12] Xu, C., Zheng, Y., Zhang, Y., Li, G., & Wang, Y. (2022). A method for detecting objects in dense scenes. *Open Computer Science*, *12*(1), 75-82.

[13] Cho, S., Paeng, J., & Kwon, J. (2022). Densely-packed object detection via hard negative-aware anchor attention. In *Proceedings of the IEEE/CVF Winter Conference on Applications of Computer Vision* (pp. 2635-2644).

[14] Hong, J., He, X., Deng, Z., & Yang, C. (2024). IoU-aware feature fusion R-CNN for dense object detection. *Machine Vision and Applications*, *35*(1), 3.

[15] Dai, L., & Liu, H. (2023, November). Deco-detr: Densely packed commodity detection with transformer. In *2023 7th Asian Conference on Artificial Intelligence Technology (ACAIT)* (pp. 453-462). IEEE.

[16] Zhong, X., Zhang, N., Hu, H., Li, L., Cen, J., & Wu, Q. (2023). Densely packed object detection with transformer-based head and EM-merger. *Service Oriented Computing and Applications*, *17*(2), 109-117.

[17] Vasanthi, P., & Mohan, L. (2023). A reliable anchor regenerative-based transformer model for x-small and dense objects recognition. *Neural Networks*, *165*, 809-829.

[18] Strohmayer, J., & Kampel, M. (2023, October). Real-time supermarket product recognition on mobile devices using scalable pipelines. In *2023 IEEE International Conference on Image Processing (ICIP)* (pp. 420-424). IEEE.

[19] Wang, C., Huang, C., Zhu, X., Li, Z., & Zhao, L. (2023). Smart retail skus checkout using improved residual network. *Scientific Reports*, *13*(1), 22512.

[20] Melek, C. G., Battini Sonmez, E., Ayral, H., & Varli, S. (2023). Development of a hybrid method for multi-stage end-to-end recognition of grocery products in shelf images. *Electronics*, *12*(17), 3640.

[21] Wang, L., Zhan, Y., Lan, L., Lin, X., Tao, D., & Gao, X. (2024). DeIoU: Toward Distinguishable Box Prediction in Densely Packed Object Detection. *IEEE Transactions on Circuits and Systems for Video Technology*, *34*(11), 11086-11100.

[22] Ghasemi Darehnaei, Z., Shokouhifar, M., Yazdanjouei, H., & Rastegar Fatemi, S. M. J. (2022). SI-EDTL: swarm intelligence ensemble deep transfer learning for multiple vehicle detection in UAV images. *Concurrency and Computation: Practice and Experience*, *34*(5), e6726.

[23] Yang, J., Shokouhifar, M., Yee, L., Khan, A. A., Awais, M., & Mousavi, Z. (2024). DT2F-TLNet: A novel text-independent writer identification and verification model using a combination of deep type-2 fuzzy architecture and Transfer Learning networks based on handwriting data. *Expert Systems with Applications*, *242*, 122704.

[24] Shokouhifar, M., & Farokhi, F. (2010, December). An artificial bee colony optimization for feature subset selection using supervised fuzzy C_means algorithm. In *3rd International conference on information security and artificial intelligent (ISAI)* (pp. 427-432).

[25] Shokouhifar, M., & Sabet, S. (2012, July). PMACO: A pheromone-mutation based ant colony optimization for traveling salesman problem. In *2012 International Symposium on Innovations in Intelligent Systems and Applications* (pp. 1-5). IEEE.

[26] Shokouhifar, M., & Jalali, A. (2014, May). Automatic symbolic simplification of analog circuits in MATLAB using ant colony optimization. In *2014 22nd Iranian Conference on Electrical Engineering (ICEE)* (pp. 407-412). IEEE.

[27] Sabet, S., Shokouhifar, M., & Farokhi, F. (2016). A comparison between swarm intelligence algorithms for routing problems. *Electrical & Computer Engineering: An International Journal (ECIJ)*, *5*(1), 17-33.

[28] Shokouhifar, M., & Jalali, A. (2016). Evolutionary based simplified symbolic PSRR analysis of analog integrated circuits. *Analog Integrated Circuits and Signal Processing*, *86*(2), 189-205.

[29] Nahavandi, B., Homayounfar, M., Daneshvar, A., & Shokouhifar, M. (2022). Hierarchical structure modelling in uncertain emergency location-routing problem using combined genetic algorithm and simulated annealing. *International Journal of Computer Applications in Technology*, *68*(2), 150-163.

[30] Shokouhifar, M., & Pilevari, N. (2022). Combined adaptive neuro-fuzzy inference system and genetic algorithm for E-learning resilience assessment during COVID-19 Pandemic. *Concurrency and Computation: Practice and Experience*, *34*(10), e6791.

[31] Hosseinzadeh, M., Yoo, J., Ali, S., Lansky, J., Mildeova, S., Yousefpoor, M. S., ... & Tightiz, L. (2023). A fuzzy logic-based secure hierarchical routing scheme using firefly algorithm in Internet of Things for healthcare. *Scientific Reports*, *13*(1), 11058.

[32] Shokouhifar, A., Shokouhifar, M., Sabbaghian, M., & Soltanian-Zadeh, H. (2023). Swarm intelligence empowered three-stage ensemble deep learning for arm volume measurement in patients with lymphedema. *Biomedical Signal Processing and Control*, *85*, 105027.

[33] Yazdanjue, N., Yazdanjouei, H., Karimianghadim, R., & Gandomi, A. H. (2024). An enhanced discrete particle swarm optimiazation for structural k-Anonymity in social networks. *Information Sciences*, *670*, 120631.

[34] Shokouhifar, M., Hasanvand, M., Moharamkhani, E., & Werner, F. (2024). Ensemble heuristic–metaheuristic feature fusion learning for heart disease diagnosis using tabular data. *Algorithms*, *17*(1), 34.

[35] Yazdanjue, N., Rakhshaninejad, M., Yazdanjouei, H., Niemelä, M. S., Chen, F., & Gandomi, A. H. (2025). Cyber threat management using semi-supervised ensemble learning and enhanced interior search algorithm: applications for illicit marketplace classification in deep/dark web and social platforms. *Annals of Operations Research*, 1-53.